\documentclass{article}
\usepackage{spconf,amsmath,graphicx}
\usepackage{subfigure}
\usepackage{amsfonts}
\usepackage{color}
\usepackage{booktabs}
\usepackage{multirow}

\DeclareMathSymbol{\shortminus}{\mathbin}{AMSa}{"39}

\newcommand{\figref}[1]{Fig. \ref{#1}}
\newcommand{\tabref}[1]{Table \ref{#1}}

\newcommand{\etal}{\textit{et al}.}
\newcommand{\ie}{\textit{i}.\textit{e}., }
\newcommand{\eg}{\textit{e}.\textit{g}. }

\title{Multi-domain Unsupervised Image-to-Image Translation\\with Appearance Adaptive Convolution}

\name{Somi Jeong$^{1\dagger}$~~~~~~~~~~~Jiyoung Lee$^{2\dagger}$~~~~~~~~~~~Kwanghoon Sohn$^{3*}$
\thanks{This research was supported by R\&D program for Advanced Integrated-intelligence for Identification (AIID) through the National Research Foundation of KOREA(NRF) funded by Ministry of Science and ICT (NRF-2018M3E3A1057289).}
\thanks{$^*$ Corresponding author $~~~~~~~^\dagger$ Work done at Yonsei University}
}
\address{{$^1$ NAVER LABS ~~~~~~~~~
$^2$ NAVER AI Lab ~~~~~~~~~$^3$ Yonsei University}\\
\tt \small {somi.jeong@naverlabs.com,~ lee.j@navercorp.com, ~khsohn@yonsei.ac.kr}}

\begin{document}
\ninept
\maketitle
\begin{abstract}
Over the past few years, image-to-image (I2I) translation methods have been proposed to translate a given image into diverse outputs.
Despite the impressive results, they mainly focus on the I2I translation between two domains, so the multi-domain I2I translation still remains a challenge.
To address this problem, we propose a novel multi-domain unsupervised image-to-image translation (MDUIT) framework that leverages the decomposed content feature and appearance adaptive convolution to translate an image into a target appearance while preserving the given geometric content.
We also exploit a contrast learning objective, which improves the disentanglement ability and effectively utilizes multi-domain image data in the training process by pairing the semantically similar images.
This allows our method to learn the diverse mappings between multiple visual domains with only a single framework.
We show that the proposed method produces visually diverse and plausible results in multiple domains compared to the state-of-the-art methods.
\end{abstract}
\section{Introduction} \vspace{-3pt}
Advances in generative adversarial networks (GANs)~\cite{goodfellow2014generative} have excelled at translating an image into a plausible and realistic image by learning a mapping between different visual domains.
More progressive unsupervised image-to-image (I2I) translation approaches~\cite{zhu2017unpaired,liu2017unsupervised} have explored learning strategies for cross-domain mapping without collecting paired data.

\begin{figure}[t!]
	\centering
	\subfigure[Multi-modal model]
	{\includegraphics[width=0.45\linewidth]{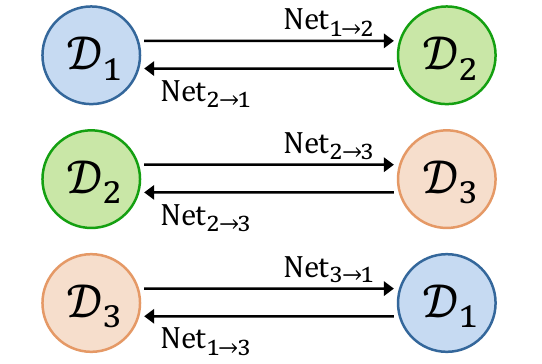}~~}
	\subfigure[Multi-domain model]
	{~~\includegraphics[width=0.45\linewidth]{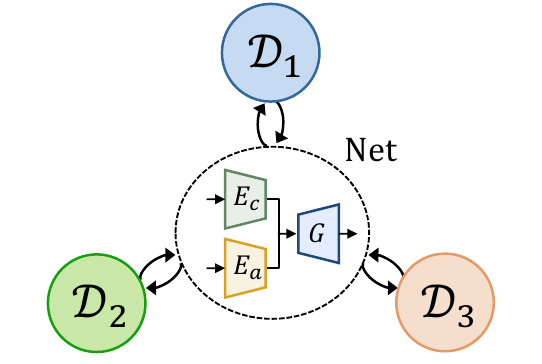}}\\
	\vspace{-8pt}
    \caption{Comparison between multi-modal and multi-domain models.
    For translating images between three domains,
    (a) the multi-modal model should train six networks respectively.
    (b) Our multi-domain model uses only a single network to deal with multiple domains.}
	\label{fig:f1} \vspace{-12pt}
\end{figure}

Recent methods~\cite{zhu2017toward,huang2018multimodal,lee2018diverse,gonzalez2018image,yang2019diversity} have presented a multi-modal I2I translation approach that produces diverse outputs in the target domain from a single input image.
It is usually formulated as a latent space disentanglement task, which decomposes the latent representation into domain-agnostic \textit{content} and domain-specific \textit{appearance}.
The content contains the intrinsic shape of objects and structures that should be preserved across domains, and the appearance contains the visually distinctive properties that are unique to each domain.
By adjusting the appearance for the fixed content, it is possible to produce visually diverse outputs while maintaining the given structure.
To achieve proper disentanglement and improve expressiveness, they apply various constraints such as weight sharing~\cite{lee2018diverse, huang2017arbitrary}, adaptive normalization~\cite{ulyanov2017improved,dumoulin2016learned,huang2017arbitrary}, and instance-wise processing~\cite{shen2019towards,bhattacharjee2020dunit,jeong2021memory}.
In that they only perform bidirectional translation between two domains, it is inefficient to translate images between diverse domains since it is necessary to train multiple generators for several cases, as shown in \figref{fig:f1} (a).

To overcome the aforementioned challenge, multi-domain I2I translation has been developed, which aims to perform simultaneous translation for all domains using only a single generative network, as illustrated in \figref{fig:f1} (b).
Thanks to its efficient handling of domain increments, attribute manipulation methods such as fashion image transformation~\cite{ak2019attribute,mo2019instagan} and facial attribute editing~\cite{choi2018stargan,lee2020maskgan,kim2019semantic} employ the multi-domain I2I translation
to transform images for various attributes.
Although they have achieved remarkable performance in translating specific local regions (\eg sleeve or mouth), they cannot yield reliable outputs for global image translation tasks such as seasonal and weather translations.
To address this issue, Yang \etal~\cite{yang2018crossing} employed two sets of encoder-decoder networks to embed features of all domains in a shared space, and Lin \etal~\cite{lin2019unsupervised} exploited multiple domain-specific decoders to generate globally translated images in multiple domains.
However, they still show poor performance when configured with very different domains.
Zhang \etal~\cite{zhang2020weakly} leveraged weakly paired images that share overlapping fields, and focused on translating high-attention parts.
Nevertheless, it may not work well when the number of domains varies in test, in that it relies on $N$-way classification loss for a fixed number of domains.

\begin{figure*}[t!]
	\centering	
	\includegraphics[width=0.9\linewidth]{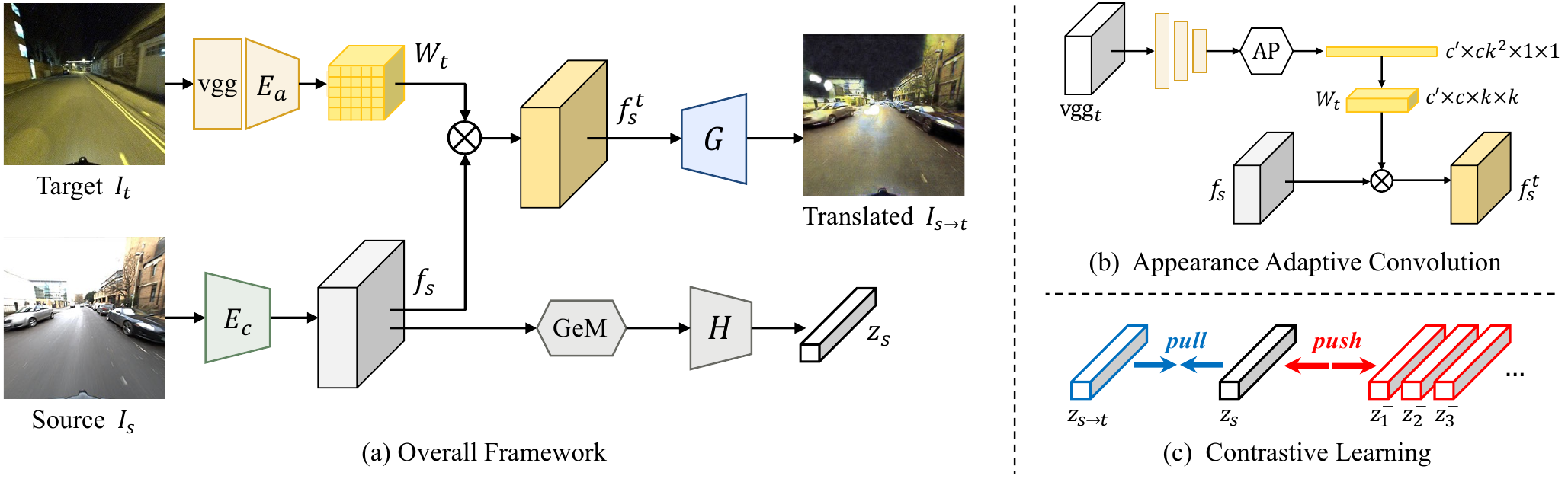}
	\vspace{-8pt}
	\caption{Illustration of (a) the proposed architecture, consisting of (b) appearance adaptive convolution and (b) contrastive learning modules.}
	\label{fig:f2}	\vspace{-8pt}
\end{figure*}

In this paper, we propose a novel multi-domain unsupervised I2I translation method (MDUIT) that learns a multi-domain mapping between various domains using only a single framework.
Similar to other I2I translation methods~\cite{zhu2017toward,huang2018multimodal,lee2018diverse}, we assume that the image representation can be decomposed into a domain-agnostic content space and domain-specific appearance space.
To flexibly translate the image into an arbitrary appearance, we opt for treating appearance as the weights of the convolution filter that is adaptively generated according to the input image.
By applying the appearance adaptive convolution on the domain-agnostic content representation, it can obtain appearance-representative content embedding, which enables multi-domain I2I translation in a unified system.
Furthermore, we leverage a contrastive learning objective, which has been a powerful tool for unsupervised visual representation learning~\cite{chen2020simple,he2020momentum}, to effectively utilize training data composed of multi-domain images while improving the ability of feature disentanglement.
Experimental results show that our method translates a given image into more diverse and realistic images than the existing methods on Oxford RobotCar dataset~\cite{maddern20171}.

\vspace{-3pt}
\section{Proposed Method}\vspace{-3pt}
Let us define by $\mathcal{D}=\{I_i\}_{i=1}^N$ consisting of images from multiple domains, where $N$ is the total number of images.
Given source and target images $\{I_s, I_t\}\in\mathcal{D}$ sampled from different domains, our goal is to produce the translated image that retains the content of $I_s$ while representing the appearance of $I_t$. 
To this end, we present the MDUIT framework consisting of two parts: 1) \emph{appearance adaptive image translation module} and 2) \emph{contrastive learning module}, as illustrated in \figref{fig:f2}.
Inspired by the dynamic filter networks~\cite{jia2016dynamic}, we introduce an appearance adaptive convolution that encodes appearance factors as filter weights to comprehensively transfer the arbitrary appearance.
In addition, we adopt a contrastive learning strategy~\cite{chen2020simple,he2020momentum} as an auxiliary task, which encourages to explicitly make semantically similar training pairs among unpaired and mixed domain image sets, as well as to better separate the domain-agnostic content representation from the image representation.

\vspace{-5pt}
\subsection{Appearance Adaptive Image Translation}\vspace{-2pt}
It aims to produce the output that analogously reflects the appearance of $I_t$ while maintaining the structural content of $I_s$.
For effective appearance propagation, we leverage the appearance adaptive convolution that stores the appearance representation in the form of convolution filter weights.
It consists of content encoder $E_c$, appearance filter encoder $E_a$, and image generator $G$.
Specifically, $I_s$ is fed into $E_c$ to extract the content feature $f_s\in \mathbb{R}^{c\times h \times w}$, which is embedded into the shared latent space across domains.
$I_t$ is fed into the pre-trained VGG-16 network~\cite{simonyan2014very}, and then the activation from `relu3-3' layer is passed to $E_a$ along with the average pooling layer.
The obtained output is reshaped to generate the appearance filter $W_t\in\mathbb{R}^{c'\times c \times k \times k}$.
In that $W_t$ is adaptively changed according to the appearance of $I_t$, it can handle any appearance translation.
The content feature $f_s$ is then aggregated to blend with the target appearance by applying $W_t$.
Finally, the appearance-representative content $f_s^t\in \mathbb{R}^{c'\times h \times w}$ is passed into $G$ and the translated image $I_{s\to t}$ is generated.
The above procedures are expressed as
\begin{equation}\label{equ:module3}
\begin{split}
f_s = E_c(I_s), ~~W_t = E_a(I_t),~~I_{s\to t} = G(f_{s}\otimes W_t),
\end{split}
\end{equation}
where $\otimes$ is 2D convolution operator.

Specifically, to transfer the target appearance into the source content, we exploit the appearance adaptive convolution, whose filter weights are driven by the target image.
We argue that the proposed appearance adaptive convolution can substitute various normalization layers such as IN~\cite{ulyanov2017improved}, CIN~\cite{dumoulin2016learned}, and AdaIN~\cite{huang2017arbitrary} that conventional image translation methods use to transfer the arbitrary appearance.
For example, these normalization layers are interpreted as a fixed $1 \times 1$ appearance filter, whose weights are first-order statistics of target appearance feature.
On the other hand, our method is a very generalized and flexible approach that allows us change the filter size and apply adaptive filter weights depending on its appearance and component properties.
As a result, it can produce a wider variety of results by directly encoding the appearance into the adaptive filter weights, and fusing them with the convolution operation.

\vspace{5pt}
\noindent \textbf{Adversarial loss. }
To further strengthen the quality of translated images, we apply the adversarial learning~\cite{goodfellow2014generative, mirza2014conditional} with image and appearance discriminators $\{D_i, D_a\}$.
Specifically, $D_i$ aims to distinguish whether the given input is a real image or a translated image, and $D_a$ aims to determine whether two concatenated images represent the same appearance or not.
To this end, we use the image and appearance adversarial losses, $\mathcal{L}_\text{adv}^{i}$ and $\mathcal{L}_\text{adv}^{a}$, as follows:
\begin{equation}\label{equ:adv}
\begin{split}
&\mathcal{L}_\text{adv}^{i}=\mathbb{E}_{I\sim \mathcal{D}}\,[\log D_i(I_s) \,\text{+} \log D_i(I_t)\, \text{+} \log (1\raisebox{1pt}{$-$} D_i(I_{s \to t}))],\\
&\mathcal{L}_\text{adv}^{a}=\mathbb{E}_{I\sim \mathcal{D}}\,[\log D_a(I_t, I_{t+})\,\text{+}\log (1\raisebox{1pt}{$-$}D_a(I_t, I_{s \to t}))],
\end{split}
\end{equation}
where $I_t$ and $I_{t+}$ are images sampled from the same domain.

\vspace{5pt}
\noindent \textbf{Reconstruction loss. }
We apply two reconstruction constraints~\cite{zhu2017unpaired,liu2017unsupervised} to force the correspondence between the input and the generated output.
It consists of two terms, self-reconstruction loss $\mathcal{L}_\text{rec}^\textit{self}$ and cycle-reconstruction loss $\mathcal{L}_\text{rec}^\textit{cyc}$, defined as
\begin{equation}\label{equ:rec}
\begin{split}
\mathcal{L}_\text{rec}^\textit{self}&=\|G\,(f_s\otimes W_s) - I_s\|_1,\\
\mathcal{L}_\text{rec}^\textit{cyc}&=\|G\,(f_{s\to t}\otimes W_s) - I_s\|_1,
\end{split}
\end{equation}
where $f_{s\to t}=E_c(I_{s \to t})$ is the content feature from $I_{s \to t}$ and $W_s=E_a(I_s)$ is the appearance filter of $I_s$.

\vspace{5pt}
\noindent \textbf{Consistency loss. }
We impose the consistency loss on the content feature $\mathcal{L}_\text{con}^c$ and the appearance filter $\mathcal{L}_\text{con}^a$ to make the components of the input and the translated image similar based on the assumption that $I_{s \to t}$ contains the same content as $I_s$ and the same appearance as $I_t$.
At the same time, we exploit the negative samples to force the content features and the appearance filters from different images to be different, resulting in better discriminative power of content and appearance.
We define the content and filter consistency losses as:
\begin{equation}\label{equ:c-content}
\begin{split}
&\mathcal{L}_\text{con}^c = \max\,(0,~ ||f_s - f_{s \to t}||_2^2 - ||f_s - f_{s-}||_2^2 + m_c),\\
&\mathcal{L}_\text{con}^a = \max\,(0,~ 1-\Delta(W_t, W_{s \to t}) + \Delta(W_t, W_{t-})),
\end{split}
\end{equation}
where $m_{c}$ is a hyper-parameter and $\Delta(x,y)$ represents the cosine similarity distance.
We consider $f_{s-}$=$E_c(I_{s-})$ and $W_{t-}$=$E_a(I_{t-})$ as negative cases, where $I_{s-}$ has the different content with $I_s$ and $I_{t-}$ is sampled from different domain with $I_t$.
These losses help to achieve the proper representation disentanglement and lead to faithful appearance control in the multi-modal image translation.

\vspace{-5pt}
\subsection{Contrastive Learning}\vspace{-2pt}
Following SimCLR~\cite{chen2020simple}, we design the contrastive learning module to maximize the similarity between the source image $I_s$ and the translated image $I_{s \to t}$ through the contrastive loss.
The main idea of contrastive learning is that the similarity of positive pairs is maximized while the similarity of negative pairs is minimized, which is intended to learn useful representation.
To this end, we sequentially append shallow networks $H$ to $E_c$ to map the content feature to the embedding space where contrastive loss is applied.
Concretely, $H$ consists of a trainable generalized-mean (GeM) pooling~\cite{radenovic2018fine}, two MLP layers, and $l_2$ normalization layer, defined as $z_* = H(f_*)$.

Specifying $I_s$ as ``query'', we define $I_{s \to t}$ as ``positive'' and randomly sampled images $\{I_i^-\}_{i=1}^{N_\text{neg}}$ as ``negative''.
These samples are mapped to a compact $K$-dimensional vector through $H$, 
and the cosine similarity between positive pair $(z_s, z_{s \to t})$ and negative pairs $\{(z_s, z_i^-)\}_{i=1}^{N_\text{neg}}$ is calculated.
We define the contrastive loss based on a noise contrastive estimation (NCE)~\cite{oord2018representation}, expressed as
\begin{equation}\label{NCE}
    \mathcal{L}_\text{NCE} = -\log \frac{\exp (z_s\cdot z_{s\to t}/\tau)}{\exp (z_s\cdot z_{s\to t}/\tau) + \sum_{i=1}^{N_\text{neg}} \exp (z_s\cdot z_i^-/\tau)},
\end{equation}
where $\tau$ is a temperature that adjusts the distances between samples.

Although $I_s$ and $I_{s\to t}$ have different appearances, this loss enforces $z_s$ and $z_{s \to t}$ to be consistent.
Therefore, it can be served as a weak supervision for training the feature disentanglement between content and appearance.
Moreover, during training, $z_s$ is used to make semantically similar source and target image pairs by comparing the similarities between training images.
If the source and target images contain disparate contents, (\eg mountain landscape in the source and urban scene in the target), the transferred appearance may hurt the translation performance and even reduce training efficiency.
In contrast, leveraging explicitly paired images offers the advantage of considering the relevant factors between them, leading to more competitive and promising results.

\vspace{-5pt}
\subsection{Full Objective}\vspace{-2pt}
To jointly train the appearance adaptive image translation and the contrastive learning modules, the final objective function to optimize $\{E_c, E_a, G, H, D_i, D_a\}$ is defined as follows:
\begin{equation}\label{equ:I2I}
\begin{split}
&\min_{E_*,G,H} \max_{D_*} \mathcal{L}(E_c, E_a, G, H, D_i, D_a)=\mathcal{L}_\text{adv}^{i} + \mathcal{L}_\text{adv}^{a} \\
&~~~+\beta_\text{rs}\mathcal{L}_\text{rec}^\textit{self} + \beta_\text{rc}\mathcal{L}_\text{rec}^\textit{cyc} + \beta_\text{cc}\mathcal{L}_\text{con}^c + \beta_\text{ca}\mathcal{L}_\text{con}^a + \beta_\text{NCE}\mathcal{L}_\text{NCE},
\end{split}
\end{equation}
where $\beta_*$ controls the relative weights between them.

\begin{figure}[t]
	\centering
	\includegraphics[width=0.95\linewidth]{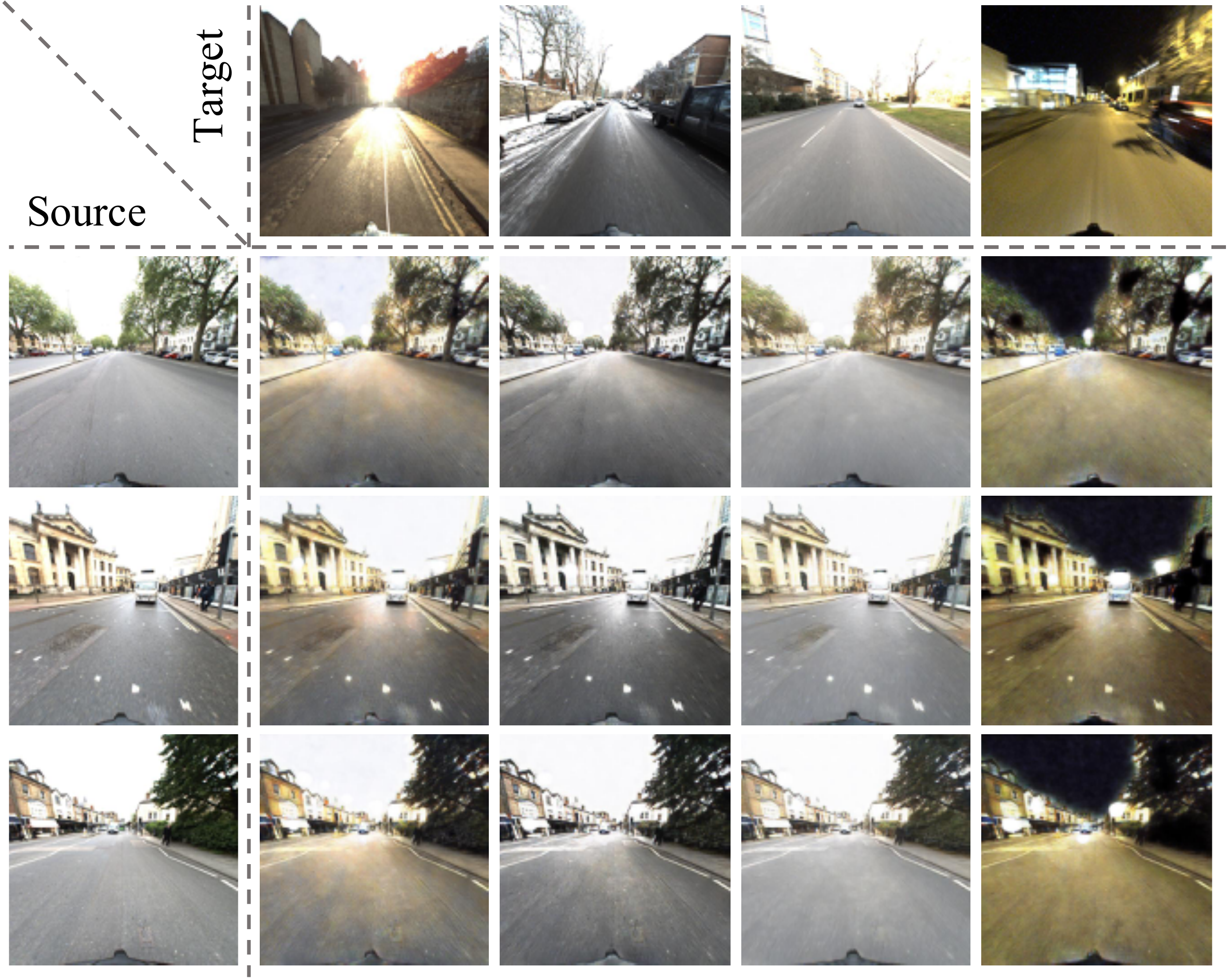}	
	\vspace{-8pt}
	\caption{Qualitative results obtained by changing source (row) and target image (column) respectively.} 
	\label{fig:f5}
	\vspace{-10pt}
\end{figure} 

\vspace{-3pt}
\section{Experiments}
\vspace{-3pt}
\subsection{Experimental Settings}
\vspace{-3pt}
\noindent \textbf{Implementation details. }
Our method was implemented based on PyTorch
and trained on NVIDIA TITAN RTX GPU.
We empirically fixed the parameters as and $\{m_c, \tau\}=\{0.1, 0.07\}$, $N_\text{neg}=16$, and $\{\beta_\text{rs}, \beta_\text{rc}, \beta_\text{cc}, \beta_\text{ca}, \beta_\text{NCE}\}=\{100, 100, 10, 1, 1\}$.
We set the appearance filter size as $\{c'\times c \times k \times k\} = \{256 \times 256 \times 5 \times 5\}$, but due to the memory capacity, we adopted group convolution, which was first introduced in AlexNet~\cite{krizhevsky2012imagenet}.
We divided the content features into 256 groups, allowing the filter size to be reduced to $1 \times 256 \times 5 \times 5 $.
The weights of all networks were initialized by a Gaussian distribution with a zero mean and a standard deviation of 0.001, and the Adam solver~\cite{kingma2015adam} was employed for optimization, where $\beta_1=0.5$, $\beta_2=0.999$, and the batch size was set to $1$.
The initial learning rate was 0.0002, kept for first 35 epochs, and linearly decayed to zero over the next 15 epochs.

\begin{figure*}[t!] 
	\centering
	\renewcommand{\thesubfigure}{}		
	\subfigure{\includegraphics[width=0.139\linewidth]{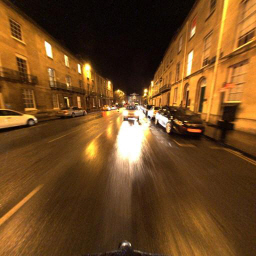}}\hfill
	\subfigure{\includegraphics[width=0.139\linewidth]{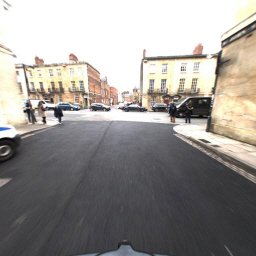}}\hfill
	\subfigure{\includegraphics[width=0.139\linewidth]{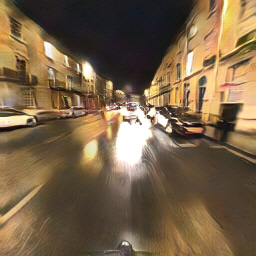}}\hfill
	\subfigure{\includegraphics[width=0.139\linewidth]{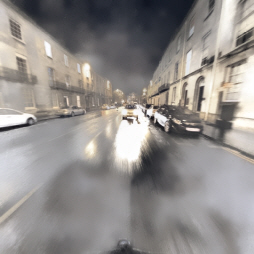}}\hfill	
	\subfigure{\includegraphics[width=0.139\linewidth]{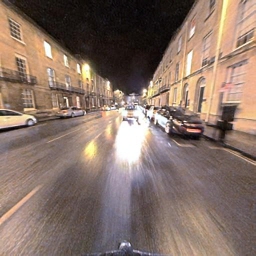}}\hfill
	\subfigure{\includegraphics[width=0.139\linewidth]{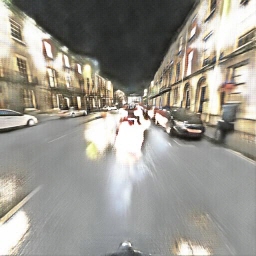}}\hfill
	\subfigure{\includegraphics[width=0.139\linewidth]{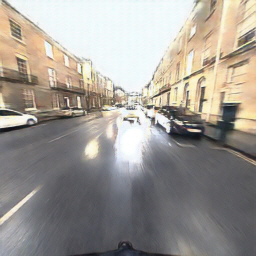}}\hfill \\
	\vspace{-9pt}
	\subfigure{\includegraphics[width=0.139\linewidth]{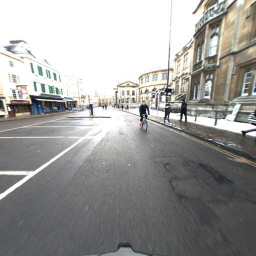}}\hfill
	\subfigure{\includegraphics[width=0.139\linewidth]{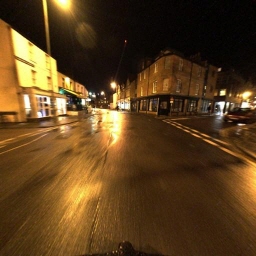}}\hfill
	\subfigure{\includegraphics[width=0.139\linewidth]{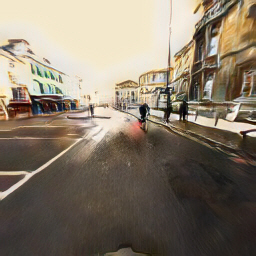}}\hfill
	\subfigure{\includegraphics[width=0.139\linewidth]{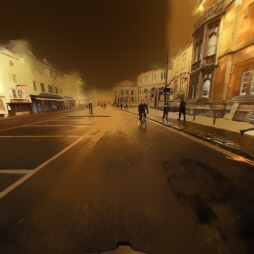}}\hfill	
	\subfigure{\includegraphics[width=0.139\linewidth]{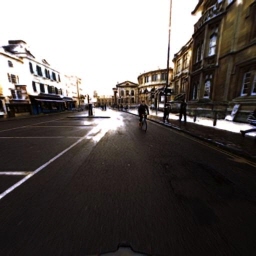}}\hfill
	\subfigure{\includegraphics[width=0.139\linewidth]{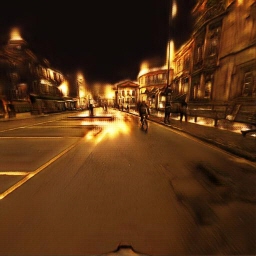}}\hfill
	\subfigure{\includegraphics[width=0.139\linewidth]{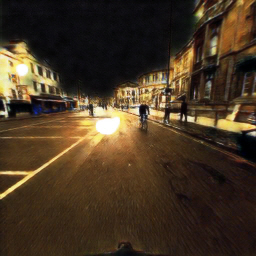}}\hfill \\
	\vspace{-9pt}
	\subfigure[(a) Source]
	{\includegraphics[width=0.139\linewidth]{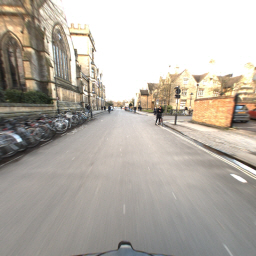}}\hfill 
	\subfigure[(b) Target]
	{\includegraphics[width=0.139\linewidth]{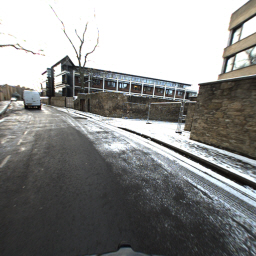}}\hfill 
	\subfigure[(c) AdaIN~\cite{huang2017arbitrary}]
	{\includegraphics[width=0.139\linewidth]{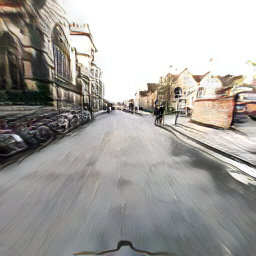}}\hfill
	\subfigure[(d) Photo-WCT~\cite{li2018closed}]
	{\includegraphics[width=0.139\linewidth]{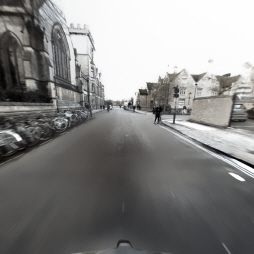}}\hfill 
	\subfigure[(e) MUNIT~\cite{huang2018multimodal}]
	{\includegraphics[width=0.139\linewidth]{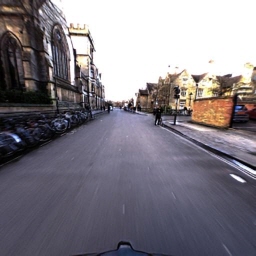}}\hfill	
	\subfigure[(f) CD-GAN~\cite{yang2018crossing}]
	{\includegraphics[width=0.139\linewidth]{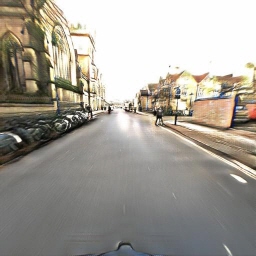}}\hfill
	\subfigure[(g) Ours]
	{\includegraphics[width=0.139\linewidth]{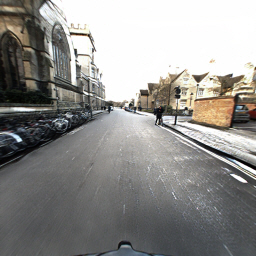}}\hfill	 \\
	\vspace{-8pt}
	\caption{Qualitative evaluations for multi-modal I2I translation on RobotCar Seasons dataset~\cite{sattler2018benchmarking}.  (Best viewed in color.)} 	\vspace{-3pt}
	\label{fig:f4}
\end{figure*}

\vspace{5pt}
\noindent \textbf{Dataset. }
We train and evaluate our model on the RobotCar Seasons dataset~\cite{sattler2018benchmarking}, which is based on publicly available Oxford Robotcar Dataset~\cite{maddern20171}.
\cite{maddern20171} is collected outdoor urban environment data on the vehicle platform over a full year, covering short-term (\emph{e.g.} time, weather) and long-term (\emph{e.g.} seasonal) changes.
The RobotCar Seasons dataset~\cite{sattler2018benchmarking} is reformed from selected images among \cite{maddern20171} to supplement their pose annotation more accurately.
It provides single appearance category (\emph{reference}) with pose annotation and the rest categories without pose annotation.

For training contrastive learning module, we built the positive and negative samples based on the pose annotation (rotation and translation).
We first forwarded the \emph{reference}-category images to the contrastive module, and calculated the similarity between $K$-dimensional vectors.
Among the most similar images, we defined the image samples with pose differences less than a threshold as positive, and otherwise as to the negative.
Here, the thresholds of rotation and translation are set as $8^{\circ}$ and $7m$.
In addition, we paired the source and target image pair for the image translation module in the same way.
We first forwarded the whole images $\mathcal{D}$ to the contrastive learning module and set the source and target images which are the most similar ones.
Not to be biased, we updated these samples every epoch.

\vspace{5pt}
\noindent \textbf{Compared methods. }
We conduct the comparison with the following SOTA methods.
AdaIN~\cite{huang2017arbitrary} and Photo-WCT~\cite{li2018closed} the arbitrary style transfer methods. 
MUNIT~\cite{huang2018multimodal} is a multi-modal I2I translation method, 
and CD-GAN~\cite{yang2018crossing} is a multi-domain I2I translation method.

\vspace{-5pt}
\subsection{Qualitative Comparison}\vspace{-3pt}
\noindent \textbf{Effect of appearance adaptive convolution. }
\figref{fig:f5} represents our translated results by changing the source image and target image respectively.
The results in the same row represent the translated images of a fixed source image with different target images, and the results in the same column represent the translated images of different source images with a fixed target image.
Although the source images are different, the translated images to the same appearance (same column) express the target appearance well, especially road and illumination.
In addition, regardless of the target appearance, the translated source images (same row) preserve their original component well.
It demonstrates superior diversity in the multi-domain image translation by generating realistic and competitive results.

\vspace{5pt}
\noindent \textbf{Comparison to State-of-the-art. }
\figref{fig:f4} shows the qualitative comparison of the state-of-the-art methods.
We observe that the arbitrary style transfer methods AdaIN~\cite{huang2017arbitrary} and Photo-WCT~\cite{li2018closed} show limited performance.
Since they simply transform the source features based on the target statistics, they tend to generate inconsistent and undesirable stylized results due to their limited feature representation capacity.
MUNIT~\cite{huang2018multimodal} and CD-GAN~\cite{yang2018crossing} show plausible results when the domain discrepancy between the source and target is not large.
However, they fail to translate image when the appearance between the source and target images is too different, \eg night and cloudy.
Compared to them, our method produces the most visually appealing images with a more vivid appearance.
We explicitly build the appearance adaptive convolution to effectively transfer the given appearance to the content features, and perform the training process between semantically similar images to improve learning efficiency, leading to superior translation fidelity as illustrated.

\vspace{-5pt}
\subsection{Multi-Domain Visual Localization}\vspace{-3pt}
To validate the discriminative capacity of $z_s$,
we applied our method to visual localization, which aims at estimating the location of an image using image retrieval.
To emphasize two important factors, \ie I2I and contrastive learning, we perform extensive evaluations \emph{with} and \emph{without} them.
As shown in \tabref{tab:1}, NetVLAD~\cite{arandjelovic2016netvlad} is the baseline to analyze relative performance, which is tailored to the visual localization.
As expected, `\emph{w/o} I2I, $\mathcal{L}_\text{NCE}$' and '\emph{w/o} I2I' show poor results because, without the I2I module\footnote{We treat a randomly color-jittered input image as the translated image.}, it is hard to obtain robust features from diverse domains.
We used a simple $L_1$ consistency loss between the positive vectors to evaluate the effect of $L_\text{NCE}$.
The performance of `\emph{w/o} I2I, $\mathcal{L}_\text{NCE}$' is clearly aggravated than '\emph{w/o} I2I' because $L_1$ loss introduces less effective representation learning.
Moreover, the result of `\emph{w/o} $L_\text{NCE}$' shows that the diversity of image appearance has a positive effect on performance, but the performance degradation is observed by replacing $L_1$ loss with $L_\text{NCE}$.
Compared to NetVLAD~\cite{arandjelovic2016netvlad}, `\emph{Full}' achieves comparable overall performance, especially better in the night domain.
It shows that our method can be further extended to solve the visual localization task under diverse visual domains.

\begin{table}[t!]
\centering
\resizebox{0.95\linewidth}{!}{
	\begin{tabular}{lcccccc}
	\toprule
	& \multicolumn{3}{c}{Day-All} & \multicolumn{3}{c}{Night-All} \\ \cmidrule(lr){2-4} \cmidrule(lr){5-7}
	\multirow{2}{*}{Method}& 0.25m & 0.5m & 5m & 0.25m & 0.5m & 5m \\
	&~2$^{\circ}$ & 5$^{\circ}$& 10$^{\circ}$~& ~2$^{\circ}$ & 5$^{\circ}$& 10$^{\circ}~$ \\ \midrule\midrule
	NetVLAD~\cite{arandjelovic2016netvlad} (baseline) & 6.4 & 26.3 & \textbf{90.9} & 0.3 & 2.3 & 15.9 \\\midrule
	\emph{w/o} I2I, $\mathcal{L}_\text{NCE}$  & 2.9 & 12.7 & 56.8 & 0.2 & 1.3 & 18.7 \\
	\emph{w/o} I2I & 5.1 & 18.9 & 74.9 & 0.7 & 4.4 & 27.7 \\
	\emph{w/o} $\mathcal{L}_\text{NCE}$ & 7.7 & 25.9 & 82.4 & 1.7 & 6.3 & 41.9 \\
	\emph{Full} &\textbf{ 8.5} & \textbf{31.8} & 88.2 & \textbf{4.1} & \textbf{14.2} & \textbf{61.0}\\  \bottomrule
    \end{tabular}}
    \vspace{-5pt}
	\caption{Quantitative comparison of visual localization on \cite{sattler2018benchmarking}.} \vspace{-10pt}
	\label{tab:1} 
\end{table}

\vspace{-4pt}
\section{Conclusion}
\vspace{-3pt}
In this paper, we present MDUIT, which translates the image into diverse appearance domains via a single framework.
Our method significantly improves the quality of the generated images compared to the state-of-the-art methods, including very challenging appearances such as night.
To this end, we propose an appearance adaptive convolution for injecting the target appearance into the content feature.
In addition, we employ contrastive learning to enhance the ability to disentangle the content and appearance features and transfer appearance between semantically similar images, thus leading to effective training.
The direction for further study is to integrate the multi-domain image translation with several interesting applications that require careful handling of multi-domain images, such as visual localization, 3D reconstruction, etc.
\newpage
\bibliographystyle{IEEEbib}
\bibliography{egbib}
\end{document}